\documentclass{article}
\PassOptionsToPackage{round,authoryear}{natbib}
\newif\ifworkshopstyle
\IfFileExists{neurips_2025.sty}{%
  \workshopstyletrue
  \usepackage[final,sglblindworkshop]{neurips_2025}
}{%
  \workshopstylefalse
  \usepackage{natbib}
  \newcommand{\workshoptitle}[1]{}
}

\workshoptitle{2nd Workshop on Principles of Generative Modeling (PriGM) at NeurIPS 2026}

\usepackage[utf8]{inputenc}
\usepackage[T1]{fontenc}
\usepackage{hyperref}
\usepackage{url}
\usepackage{booktabs}
\usepackage{amsmath,amssymb,amsfonts,amsthm}
\usepackage{mathtools}
\usepackage{bm}
\usepackage{nicefrac}
\usepackage{microtype}
\usepackage[dvipsnames]{xcolor}
\usepackage{graphicx}
\usepackage{subcaption}
\usepackage{wrapfig}

\ifworkshopstyle\else
  \def\And{\end{tabular}\hfil\linebreak[0]\hfil
    \begin{tabular}[t]{c}\bfseries\ignorespaces}
  \def\AND{\end{tabular}\hfil\linebreak[4]\hfil
    \begin{tabular}[t]{c}\bfseries\ignorespaces}
\fi

\newcommand{\preprintgraphic}[2][]{%
  \IfFileExists{#2}{\includegraphics[#1]{#2}}{%
    \fbox{\parbox[c][34mm][c]{0.9\linewidth}{\centering\small
      Figure asset: \texttt{\detokenize{#2}}}}}}

\newcommand{\E}{\mathbb{E}}
\newcommand{\R}{\mathbb{R}}
\newcommand{\cN}{\mathcal{N}}
\newcommand{\task}{\bm{w}}
\newcommand{\x}{\bm{x}}
\newcommand{\q}{\bm{x}_{\mathrm{test}}}
\newcommand{\e}{\bm{\varepsilon}}

\newcommand{\vv}[1]{\mathrm{vec}\!\left(#1\right)}
\newcommand{\iid}{\stackrel{\mathrm{i.i.d.}}{\sim}}
\newcommand{\tr}{\operatorname{tr}}

\usepackage[most]{tcolorbox}
\definecolor{SideBarGreen}{HTML}{1A4B31}
\definecolor{TextDarkGreen}{HTML}{1A4B31}
\definecolor{GreenBorder}{HTML}{A0B8A9}
\definecolor{GreenPanel}{HTML}{F5FAF7}
\definecolor{SideBarBrown}{HTML}{613D04}
\definecolor{TextDarkBrown}{HTML}{362205}
\definecolor{BrownBorder}{HTML}{C2B8A5}
\definecolor{BrownPanel}{HTML}{F7F5F2}
\definecolor{CiteColor}{HTML}{4065A1}
\definecolor{SideBarBlue}{HTML}{2E4770}
\definecolor{TextDarkBlue}{HTML}{2E4770}
\definecolor{BlueBorder}{HTML}{A9BDD8}
\definecolor{BluePanel}{HTML}{F5F8FC}
\tcbset{
  resultbox/.style={
    enhanced,
    breakable,
    sharp corners,
    boxrule=0.55pt,
    opacityback=1,
    boxsep=0pt,
    left=11pt,
    right=11pt,
    top=9pt,
    bottom=9pt,
    before skip=10pt,
    after skip=10pt,
    fonttitle=\bfseries,
    detach title,
  },
  greenpalette/.style={
    colframe=GreenBorder,
    colback=GreenPanel,
    borderline west={1.6pt}{0pt}{SideBarGreen},
    coltitle=TextDarkGreen,
    before upper={
      \noindent{\color{SideBarGreen}\bfseries\tcbtitle}\par
      \vspace{0.35em}
      {\color{GreenBorder!45}\hrule height 0.35pt}
      \vspace{0.75em}
    },
  },
  brownpalette/.style={
    colframe=BrownBorder,
    colback=BrownPanel,
    borderline west={1.6pt}{0pt}{SideBarBrown},
    coltitle=TextDarkBrown,
    before upper={
      \noindent{\color{SideBarBrown}\bfseries\tcbtitle}\par
      \vspace{0.35em}
      {\color{BrownBorder!45}\hrule height 0.35pt}
      \vspace{0.75em}
    },
  },
  bluepalette/.style={
    colframe=BlueBorder,
    colback=BluePanel,
    borderline west={1.6pt}{0pt}{SideBarBlue},
    coltitle=TextDarkBlue,
    before upper={
      \noindent{\color{SideBarBlue}\bfseries\tcbtitle}\par
      \vspace{0.35em}
      {\color{BlueBorder!45}\hrule height 0.35pt}
      \vspace{0.75em}
    },
  },
}

\hypersetup{
  colorlinks=true,
  linkcolor=CiteColor,
  citecolor=CiteColor,
  urlcolor=CiteColor
}

\newenvironment{boxedproposition}[1]{%
  \refstepcounter{proposition}%
  \begin{tcolorbox}[
    resultbox,
    bluepalette,
    title={Result~\theproposition: #1}
  ]%
}{%
  \end{tcolorbox}%
}
\newenvironment{boxedlemma}[1]{%
  \refstepcounter{lemma}%
  \begin{tcolorbox}[
    resultbox,
    bluepalette,
    title={Lemma~\thelemma: #1}
  ]%
}{%
  \end{tcolorbox}%
}

\newcommand{\unmarkedfootnote}[1]{%
  \begingroup
  \renewcommand{\thefootnote}{}%
  \footnote{#1}%
  \addtocounter{footnote}{-1}%
  \endgroup
}

\title{What can linear attention learn \\ from nonlinear teachers in-context?}

\author{%
  Mary Letey\thanks{Equal contribution.} \\
  Applied Mathematics, Harvard University \\
  Kempner Institute, Harvard University \\
  \texttt{\textcolor{CiteColor}{maryletey@fas.harvard.edu}}
  \And
  Arman Rysmakhanov\footnotemark[1]%
  \;\;\thanks{This work was performed while at Harvard University.} \\
  Williams College \\
  Applied Mathematics, Harvard University \\
  \textcolor{CiteColor}{\texttt{ar21@williams.edu}}
  \AND
  Yue M.~Lu \\
  Applied Mathematics, Harvard University \\
  \textcolor{CiteColor}{\texttt{yuelu@seas.harvard.edu}}
  \And
  Cengiz Pehlevan \\
  Applied Mathematics, Harvard University \\
  Kempner Institute, Harvard University \\
  \textcolor{CiteColor}{\texttt{cpehlevan@seas.harvard.edu}}
  \And
  Jacob A.~Zavatone-Veth \\
  Center for Brain Science, Harvard University \\
  Society of Fellows, Harvard University \\
  \textcolor{CiteColor}{\texttt{jzavatoneveth@fas.harvard.edu}}
}
\date{}

\begin{document}
\maketitle
\unmarkedfootnote{%
  Code for figures is available on
  \href{https://github.com/mletey/single-index-icl}{{github}}.
}

\begin{abstract}
Linear attention is a tractable model for understanding the mechanisms governing in-context learning in transformers. For linear regression tasks, recent asymptotic analyses have characterised its learning and generalisation behaviour. We extend this theory to nonlinear single-index targets, $y=f(\x^\top \task)+\varepsilon$. Our main result establishes a nonlinearity--noise equivalence: linear attention extracts only the linear Hermite component of $f$, while the remaining nonlinear structure contributes to the generalisation error as effective noise. This reduction allows results from the corresponding linear theory to be transferred to nonlinear tasks. We illustrate its implications for finite pretraining data and for the transition from task memorisation to task generalisation as task diversity increases. These results identify a limitation of the reduced linear-attention model and provide a tractable starting point for studying nonlinear in-context learning.
\end{abstract}

\section{Introduction}

Linear attention is a useful setting for studying in-context learning because its simplified structure admits detailed analysis of the computations performed over prompt data. This led to work connecting in-context learning to regression and statistical estimation, including recent asymptotic analyses that describe generalisation in linear-regression tasks \citep{zhang2023doesincontextlearninglearn,vonoswald2023transformers,lu2025asymptotictheoryincontextlearning, bordelon2025theoryscalinglawsincontext, letey2025pretrain, letey2026sequential}.

At the same time, linear-attention mechanisms are becoming more prevalent in real-world models. Earlier work focused on computational efficiency and recurrent fast-weight interpretations \citep{katharopoulos2020transformers,schlag2021linear}, while more recent architectures introduce delta-rule updates and gating to improve recurrent memory \citep{yang2024parallelizing,yang2023gated}. This motivates the study of in-context learning using linear attention; specifically, we must ask which components of a nonlinear task a linear-attention block can extract in-context, and which components must instead be supplied by the surrounding architecture.

We investigate an extension of the linear-regression ICL model induced by a reduced linear-attention block \citep{lu2025asymptotictheoryincontextlearning,zhang2023doesincontextlearninglearn}. The key extension is the addition of nonlinear target functions, \textit{i.e.}, single-index targets. This is a classic extension of regression in statistical estimation \citep{mei2022generalization,gerace2020generalisation}, appearing in both the standard supervised-learning literature and more recent ICL results. Concurrent work by \cite{gu2026incontextlearningsingleindextargets} also studies this problem, with the addition of nonlinear readout on linear attention, thus adding expressivity to learn beyond the linear mode. Alternatively, we focus specifically on linear attention and highlight its limitations on this task. \citet{oko2024pretrainedtransformerefficientlylearns} show that transformers learn single-index tasks in context at a rate governed by the information exponent of the link function, defined as the index of its first nonzero Hermite coefficient. Related dependence on the information exponent appears in non-ICL settings \citep{damian2023smoothing}. We likewise analyse the link function by expanding it in Hermite polynomials \citep{joshi2026learning,eftekhari2021inference,dudeja2018learning}.

This work serves to formalise the following intuition: linear attention cannot learn beyond the linear Hermite mode of the link function. If no linear mode is present, all signal is indistinguishable from noise. This hypothesis is natural from standard results studying learnability of single-index targets \citep{li1989regression,brillinger2011generalized,thrampoulidis2015lasso}: when the student and link function are not compatible the full target nonlinearity cannot be learned. We expect this to apply to our case of linear attention here, as due to the lack of explicit nonlinear activation in the linear attention architecture, it cannot represent nonlinear functions of the target label. This is precisely what we show in the following. 

\section{Setup}

\paragraph{Single-index tasks.} The model is trained across regression tasks and must use the examples in a prompt to predict the response to a new covariate. 

A prompt is a sequence
\[
\{\x_1,y_1,\ldots,\x_\ell,y_\ell,\q\},
\]
with
\begin{equation}
  y_i=f(\x_i^\top\task)+\varepsilon_i,
  \qquad
  y_{\mathrm{test}}=f(\q^\top\task)+\varepsilon_{\mathrm{test}}.
  \label{eq:single-index}
\end{equation}
We take $\x_i,\q\iid\cN(\bm{0},I_d/d)$ and independent Gaussian label noise
$\e\sim\cN(\bm{0},\rho I_\ell)$ and
$\varepsilon_{\mathrm{test}}\sim\cN(0,\rho)$. The parameters are set from $n$ such samples of these sequences, each with a potentially different defining task $\task$. As in prior work, we consider two distributions over tasks. For the training sequences, we use $k$ tasks sampled independently from $\mathcal{N}(\bm{0},C_\mathrm{train})$; these $k$ tasks are split uniformly across the $n$ sequences in the training batch. At test time, the model is tested on sequences with tasks sampled independently from $\mathcal{N}(\bm{0}, C_\mathrm{test}).$

\paragraph{Nonlinearity.}
For $g\sim\cN(0,1)$, expand $f$ in an orthonormal Hermite basis:
\begin{equation}
  f(x)=\sum_{m=0}^{\infty}a_mh_m(x),
  \qquad
  a_m=\E_{g\sim\cN(0,1)}[f(g)h_m(g)].
  \label{eq:hermite}
\end{equation}
We assume $f$ has at most polynomial growth, i.e., $|f(x)| < A (1 + |x|^s)$ for some power $s$ and constant $A$.

\paragraph{Proportional limit.}
We take $d\to\infty$ with
\begin{equation}
  \kappa=\frac{k}{d},
  \qquad
  \alpha=\frac{\ell}{d},
  \qquad
  \tau=\frac{n}{d^2},
  \label{eq:prop_limit_vars}
\end{equation}
all of order one, as in prior work \citep{lu2025asymptotictheoryincontextlearning,letey2025pretrain}.

\paragraph{Reduced linear-attention predictor.}
The next-token prediction is approximated by
\begin{equation}
  \widehat y_{\mathrm{test}}=\langle\Gamma,H\rangle_F,
  \qquad
  \Gamma\in\R^{d\times(d+1)},
  \label{eq:prediction}
\end{equation}
where, for one prompt,
\begin{equation}
  H=\x_{\mathrm{test}}
  \begin{bmatrix}
    \displaystyle\frac d\ell\sum_{i=1}^{\ell}y_i\x_i^\top
    &
    \displaystyle\frac1\ell\sum_{i=1}^{\ell}y_i^2
  \end{bmatrix}
  \in\R^{d\times(d+1)}.
  \label{eq:H_def}
\end{equation}
For $n$ training prompts, write $\bm{\gamma}=\vv{\Gamma}$ and $\bm{h}_\mu=\vv{H^\mu}$. Ridge regression gives
\begin{equation}
  \bm{\gamma}^*
  =
  \left(
    \frac nd\lambda I_{d(d+1)}
    +\sum_{\mu=1}^n\bm{h}_\mu\bm{h}_\mu^\top
  \right)^{-1}
  \sum_{\mu=1}^ny_{\mathrm{test}}^\mu\bm{h}_\mu,
  \label{eq:gamma_definition}
\end{equation}
and the test error is
\begin{equation}
  \mathcal E_{\mathrm{ICL}}(\Gamma^*)
  =
  \E_{\mathrm{new}}
  \left[
    \left(
      y_{\mathrm{test}}^{\mathrm{new}}
      -\langle H^{\mathrm{new}},\Gamma^*\rangle_F
    \right)^2
  \right].
  \label{eq:e_of_gamma_random_definition}
\end{equation}

\section{Theoretical results and analysis}
Define $\rho_f$ and $\rho_\mathrm{nonlinear}$ as 
\begin{equation}
  \rho_f:=\rho+\sum_{m=0}^{\infty}a_m^2 = \left(\rho + \sum_{m\neq 1} a_m^2\right) + a_1^2 := \rho_\mathrm{nonlinear} + a_1^2
  \label{eq:rhof},
\end{equation}
where $\rho$ is the label-noise variance.

\begin{boxedproposition}{Asymptotic formula for ridgeless ICL error}\label{prop:main_formula}
Suppose that the pretraining and test tasks are sampled from $\mathcal{N}(\bm{0},C_\mathrm{train})$ and $\mathcal{N}(\bm{0},C_\mathrm{test})$, respectively, where
\begin{align*}
    \tr(C_\mathrm{train}), \tr(C_\mathrm{test}) &= d,\\
    \tr(C^2_\mathrm{train}), \tr(C^2_\mathrm{test}) &= o(d^2). 
\end{align*}
If the link function $f$ has a nonzero linear mode, $a_1 \neq 0$, then
\begin{equation}
    e_\mathrm{ICL}^\mathrm{nonlinear}(C_\mathrm{train}, C_\mathrm{test}, \rho) = e_\mathrm{ICL}^\mathrm{linear}(a_1^2C_\mathrm{train}, a_1^2C_\mathrm{test}, \rho_\mathrm{nonlinear}).
\end{equation}
If instead $a_1=0$, then
\begin{equation}
  e^\mathrm{nonlinear}_{\mathrm{ICL}}
  =
  \rho_\mathrm{nonlinear}
  \begin{cases}
    \displaystyle\frac1{1-\tau},&\tau<1,\\[2mm]
    \displaystyle\frac{\tau}{\tau-1},&\tau>1.
  \end{cases}
  \label{eq:no-linear-mode}
\end{equation}
\end{boxedproposition}

The derivation is given in Appendix~\ref{sec:icl_formula_proof}. In other words, linear attention learns the linear mode, while the other modes appear as noise. If there is no linear mode to learn, ICL reduces to no-signal regression with sample ratio $\tau = n/d^2$ and label noise $\rho_\mathrm{nonlinear}$. In this case, the error has no dependence on context length or task diversity, since those quantities modulate the learnability of a linear component that is not present.

\begin{figure}[htbp]
    \centering
    \preprintgraphic[width=0.7\linewidth]{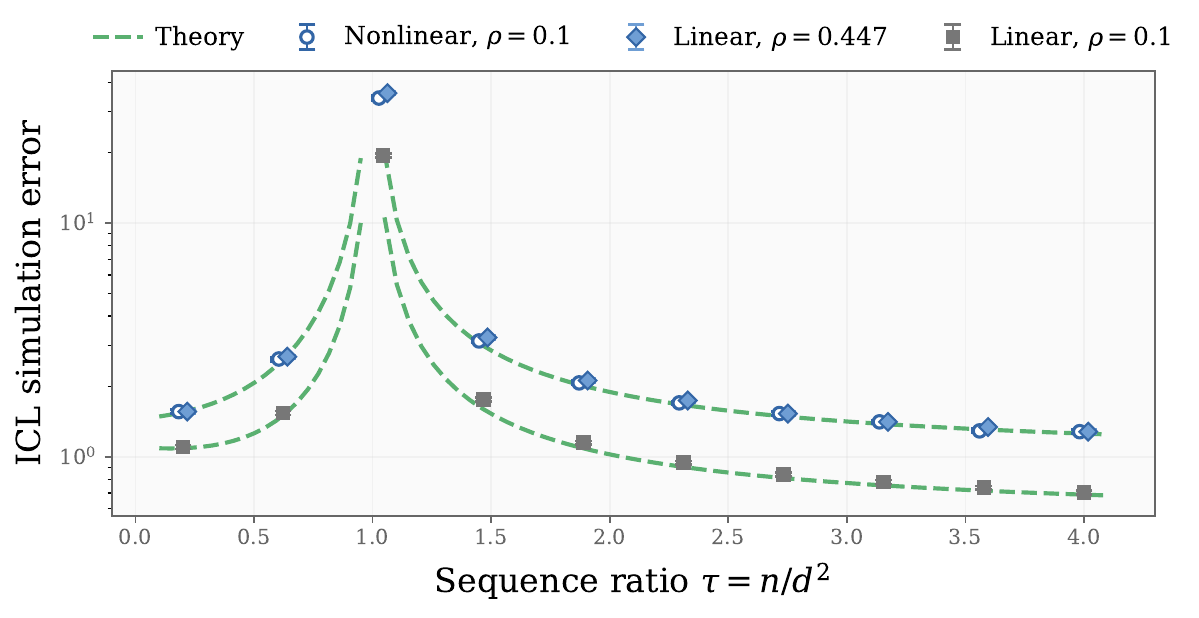}
    \caption{We simulate ICL test loss directly from data to verify Result \ref{prop:main_formula}. The linear target is $y = \task^\top \x$, while the nonlinear target is $\sin(\task^\top\x)$. $d = 64$, $\alpha = 2.0$, $\kappa = 2.0$, $C_\mathrm{train} = I_d = C_\mathrm{test}$.}
    \label{fig:loss_vs_tau}
\end{figure}

Figure \ref{fig:loss_vs_tau} provides a demonstration of Result \ref{prop:main_formula}, particularly the change in effective noise. The ICL performance of the nonlinear function at the baseline noise level is therefore empirically equivalent to that of the linear function at a higher noise level. This new effective noise is computed as $(a_1^2 + \rho_\mathrm{nonlinear})/a_1^2$ to balance the $a_1^2$ scaling of the task covariances in Result \ref{prop:main_formula}.

\paragraph{Phase transition in task diversity.}
Previous work has found that taking $\alpha,\tau\to\infty$ with $\tau\propto\alpha$ gives a sharp transition at $\kappa= \text{rank}(C_\mathrm{train})/d$. This corresponds to a transition between the model memorising the pretraining tasks and actually generalising to the correct task structure for good performance on $C_\text{test}$.

\begin{figure}[htbp]
    \centering
    \preprintgraphic[width=0.55\linewidth]{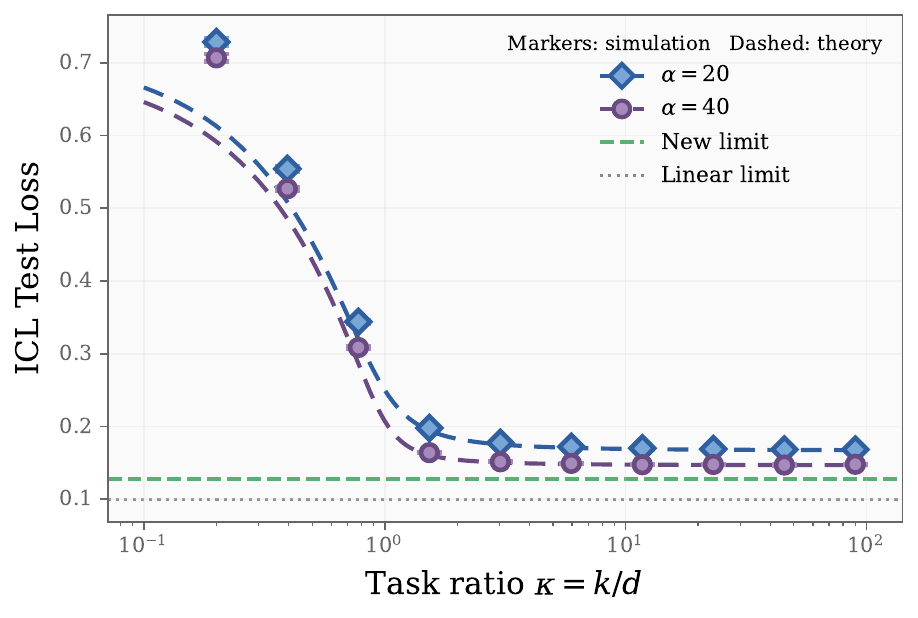}
    \caption{We see that nonlinear functions raise the limiting ICL test error, even in the best possible data case of $\alpha,\kappa,\tau \to \infty$ due to nonlearnable signal captured as noise. Here $d=32, \tau=\alpha/2,\rho=0.1$, $f(x) = \tanh(x).$}
    \label{fig:phase}
\end{figure}

Indeed, if we take $C_\mathrm{train} = C_\mathrm{test} = I_d$ for simplicity, when $\kappa>1$ the error approaches the new noise floor (so long as $a_1\neq 0$)
\begin{equation}
  e_{\mathrm{ICL}}(\Gamma^*)
  \longrightarrow \rho_\text{nonlinear}
  \label{eq:task-diversity-limit}
\end{equation}
rather than the noise floor $\rho$ obtained for a linear target \citep{lu2025asymptotictheoryincontextlearning}. This is demonstrated in Figure \ref{fig:phase}.

\section{Conclusion}
We have extended the asymptotic theory of in-context learning by a reduced linear-attention block from linear regression tasks to nonlinear single-index targets. The resulting equivalence shows that the model uses the target function only through its linear Hermite component, while the remaining Hermite modes contribute as effective noise. Consequently, increasing the amount of pretraining data or the diversity of pretraining tasks cannot remove the error floor created by nonlinear components that are invisible to this predictor. This limitation persists beyond the task-diversity transition, where the error approaches $\rho_\mathrm{nonlinear}$ rather than the original label-noise level $\rho$. 

\begin{ack}
    M.L. is supported by a Graduate Fellowship from the Kempner Institute for the Study of Natural and Artificial Intelligence. A.R. was supported by the Alumni Sponsored Internship Program (ASIP) of Williams College. Y.M.L. gratefully acknowledges support from a Harvard College Professorship, the Harvard FAS Dean’s Fund for Promising Scholarship, and DARPA grant DIAL-FP-038. C.P. is supported by an NSF CAREER Award (IIS-2239780), DARPA grants DIAL-FP-038 and AIQ-HR00112520041, and the Simons Collaboration on the Physics of Learning and Neural Computation. J.Z.-V. was supported by a Junior Fellowship from the Harvard Society of Fellows. This work has been made possible in part by a gift from the Chan Zuckerberg Initiative Foundation to establish the Kempner Institute for the Study of Natural and Artificial Intelligence.

\end{ack}

\IfFileExists{references.bib}{%
  \bibliographystyle{plainnat}
  \bibliography{references}

@inproceedings{schlag2021linear,
  title={Linear transformers are secretly fast weight programmers},
  author={Schlag, Imanol and Irie, Kazuki and Schmidhuber, J{\"u}rgen},
  booktitle={International conference on machine learning},
  pages={9355--9366},
  year={2021},
  organization={PMLR}
}

@article{yang2024parallelizing,
  title={Parallelizing linear transformers with the delta rule over sequence length},
  author={Yang, Songlin and Wang, Bailin and Zhang, Yu and Shen, Yikang and Kim, Yoon},
  journal={Advances in neural information processing systems},
  volume={37},
  pages={115491--115522},
  year={2024}
}

@misc{gu2026incontextlearningsingleindextargets,
      title={In-context Learning of Single-index Targets: Comparing Kernel and Feature Learners}, 
      author={Haotian Gu and Yizhou Xu and Lenka Zdeborová},
      year={2026},
      eprint={2610.01712},
      archivePrefix={arXiv},
      primaryClass={cs.LG},
      url={https://arxiv.org/abs/2610.01712}, 
}

@article{yang2023gated,
  title={Gated linear attention transformers with hardware-efficient training},
  author={Yang, Songlin and Wang, Bailin and Shen, Yikang and Panda, Rameswar and Kim, Yoon},
  journal={arXiv preprint arXiv:2312.06635},
  year={2023}
}

@misc{bordelon2025theoryscalinglawsincontext,
	title        = {Theory of Scaling Laws for In-Context Regression: Depth, Width, Context and Time},
	author       = {Blake Bordelon and Mary I. Letey and Cengiz Pehlevan},
	year         = {2025},
	url          = {https://arxiv.org/abs/2510.01098},
	eprint       = {2510.01098},
	archiveprefix = {arXiv},
	primaryclass = {stat.ML}
}

@inproceedings{gerace2020generalisation,
	title        = {Generalisation error in learning with random features and the hidden manifold model},
	author       = {Gerace, Federica and Loureiro, Bruno and Krzakala, Florent and M{\'e}zard, Marc and Zdeborov{\'a}, Lenka},
	year         = {2020},
	booktitle    = {International Conference on Machine Learning},
	pages        = {3452--3462},
	organization = {PMLR}
}

@inproceedings{katharopoulos2020transformers,
	title        = {Transformers are {RNN}s: Fast autoregressive transformers with linear attention},
	author       = {Katharopoulos, Angelos and Vyas, Apoorv and Pappas, Nikolaos and Fleuret, Fran{\c{c}}ois},
	year         = {2020},
	booktitle    = {International conference on machine learning},
	pages        = {5156--5165},
	organization = {PMLR}
}

@inproceedings{letey2025pretrain,
	title        = {Pretrain-Test Task Alignment Governs Generalization in In-Context Learning},
	author       = {Mary Letey and Jacob A Zavatone-Veth and Yue M. Lu and Cengiz Pehlevan},
	year         = {2026},
	booktitle    = {The Fourteenth International Conference on Learning Representations},
	url          = {https://openreview.net/forum?id=KZLeg0MQ2r}
}

@inproceedings{letey2026sequential,
	title        = {Sequential Correlations Change In-Context Learning: Effective Context Length and Architectural Mismatch},
	author       = {Mary Letey and Yue M. Lu and Cengiz Pehlevan and Jacob A Zavatone-Veth},
	year         = {2026},
	booktitle    = {High-dimensional Learning Dynamics 2026},
	url          = {https://openreview.net/forum?id=AYprbQxAbI}
}

@article{lu2025asymptotictheoryincontextlearning,
	title        = {Asymptotic theory of in-context learning by linear attention},
	author       = {Yue M. Lu  and Mary Letey  and Jacob A. Zavatone-Veth  and Anindita Maiti  and Cengiz Pehlevan},
	year         = {2025},
	journal      = {Proceedings of the National Academy of Sciences},
	volume       = {122},
	number       = {28},
	pages        = {e2502599122},
	doi          = {10.1073/pnas.2502599122},
	url          = {https://www.pnas.org/doi/abs/10.1073/pnas.2502599122},
	eprint       = {https://www.pnas.org/doi/pdf/10.1073/pnas.2502599122}
}

@article{mei2022generalization,
	title        = {The generalization error of random features regression: Precise asymptotics and the double descent curve},
	author       = {Mei, Song and Montanari, Andrea},
	year         = {2022},
	journal      = {Communications on Pure and Applied Mathematics},
	publisher    = {Wiley Online Library},
	volume       = {75},
	number       = {4},
	pages        = {667--766}
}

@misc{oko2024pretrainedtransformerefficientlylearns,
	title        = {Pretrained transformer efficiently learns low-dimensional target functions in-context},
	author       = {Kazusato Oko and Yujin Song and Taiji Suzuki and Denny Wu},
	year         = {2024},
	url          = {https://arxiv.org/abs/2411.02544},
	eprint       = {2411.02544},
	archiveprefix = {arXiv},
	primaryclass = {cs.LG}
}

@article{damian2023smoothing,
  title={Smoothing the landscape boosts the signal for sgd: Optimal sample complexity for learning single index models},
  author={Damian, Alex and Nichani, Eshaan and Ge, Rong and Lee, Jason D},
  journal={Advances in Neural Information Processing Systems},
  volume={36},
  pages={752--784},
  year={2023}
}

@article{joshi2026learning,
  title={Learning single index models via harmonic decomposition},
  author={Joshi, Nirmit and Koubbi, Hugo and Misiakiewicz, Theodor and Srebro, Nati},
  journal={Advances in Neural Information Processing Systems},
  volume={38},
  pages={45052--45127},
  year={2026}
}

@article{eftekhari2021inference,
  title={Inference in high-dimensional single-index models under symmetric designs},
  author={Eftekhari, Hamid and Banerjee, Moulinath and Ritov, Ya'acov},
  journal={Journal of Machine Learning Research},
  volume={22},
  number={27},
  pages={1--63},
  year={2021}
}

@inproceedings{dudeja2018learning,
  title={Learning single-index models in gaussian space},
  author={Dudeja, Rishabh and Hsu, Daniel},
  booktitle={Conference On Learning Theory},
  pages={1887--1930},
  year={2018},
  organization={PMLR}
}

@article{thrampoulidis2015lasso,
  title={Lasso with non-linear measurements is equivalent to one with linear measurements},
  author={Thrampoulidis, Christos and Abbasi, Ehsan and Hassibi, Babak},
  journal={Advances in Neural Information Processing Systems},
  volume={28},
  year={2015}
}

@article{li1989regression,
  title={Regression analysis under link violation},
  author={Li, Ker-Chau and Duan, Naihua},
  journal={The Annals of Statistics},
  volume={17},
  number={3},
  pages={1009--1052},
  year={1989},
  publisher={Institute of Mathematical Statistics}
}

@incollection{brillinger2011generalized,
  title={A generalized linear model with “Gaussian” regressor variables},
  author={Brillinger, David R},
  booktitle={Selected Works of David Brillinger},
  pages={589--606},
  year={2011},
  publisher={Springer}
}

@inproceedings{vonoswald2023transformers,
	title        = {Transformers Learn In-Context by Gradient Descent},
	author       = {Von Oswald, Johannes and Niklasson, Eyvind and Randazzo, Ettore and Sacramento, João and Mordvintsev, Alexander and Zhmoginov, Andrey and Vladymyrov, Max},
	year         = {2023},
	month        = {23--29 Jul},
	booktitle    = {Proceedings of the 40th International Conference on Machine Learning},
	publisher    = {PMLR},
	series       = {Proceedings of Machine Learning Research},
	volume       = {202},
	pages        = {35151--35174},
	url          = {https://proceedings.mlr.press/v202/von-oswald23a.html},
	editor       = {Krause, Andreas and Brunskill, Emma and Cho, Kyunghyun and Engelhardt, Barbara and Sabato, Sivan and Scarlett, Jonathan}
}

@misc{zhang2023doesincontextlearninglearn,
	title        = {What and How does In-Context Learning Learn? Bayesian Model Averaging, Parameterization, and Generalization},
	author       = {Yufeng Zhang and Fengzhuo Zhang and Zhuoran Yang and Zhaoran Wang},
	year         = {2023},
	url          = {https://arxiv.org/abs/2305.19420},
	eprint       = {2305.19420},
	archiveprefix = {arXiv},
	primaryclass = {stat.ML}
}
}{%
  \begin{center}\small\color{CiteColor}
  Bibliography omitted in this single-file preview. The project build uses
  \texttt{references.bib}.
  \end{center}
}

\clearpage
\appendix

\section[Gaussian equivalent for xw]{Gaussian equivalent for $\task^\top \x$}\label{sec:gaussian_universality_general_covariance}
The input to the nonlinearity will be $\task^\top \x$. In order to expand in the basis of Hermite polynomials, we must first understand the distribution of this random variable. We know that, conditional on fixed task $\task$, $$\task^\top \x \overset{d}= \sqrt{\sigma_{\task}}g$$ for $$\sigma_{\task} = \frac{1}{d}\|\task\|^2, \qquad g \sim \mathcal{N}(0,1) \text{ independent of $\x, \task$.}$$

Now consider the distribution of the scale random variable $\sigma_{\task}$ induced by $\task$. Let $\task \sim \mathcal{N}(\bm{0}, C)$ for $C \in \mathbb{R}^{d\times d}$. For a $d \times d$ matrix $A$, define $\tr[A] := \tr(A)/d$. We have that $$\E[\sigma_w] = \tr[C],
\qquad
\mathrm{Var}(\sigma_w)=2\frac{\tr(C^2)}{d^2}.$$
We will thus assume two conditions on $C$. \begin{itemize}
    \item \textbf{Scaling freedom}. We fix $\tr(C) = d$ or $\tr[C] = 1$. If the goal is to study the effect of isotropy or anisotropy induced by task covariance $C$, it makes sense to normalise to total amount of signal, which is equivalent to normalising $\tr(C)$. This also corresponds to normalising the strength of the input to the nonlinearity, which is also standard in actual neural networks. 
    \item \textbf{Sufficient effective rank}. We will of course need to control the fluctuations of the scale $\sigma_{\task}$ away from 1. This corresponds to controlling the effective rank: we will assume $$\frac{\tr(C^2)}{d^2} \to 0 \,\; \mathrm{ as } \; d\to \infty.$$ Note that this is not a superfluous condition; it is precisely the condition that will be needed for the later application of the formulas from \cite{letey2025pretrain}.
\end{itemize} 

To justify representing $\task^\top \x$ by $g$ in the moments below, first consider the fourth moment
\[
\E_g[f^4(\sqrt{\sigma_{\task}} g) \, | \, \sigma_w]
=
\frac{1}{\sqrt{2\pi\sigma_{\task}}}
\int_{\mathbb R}
f^4(z)\exp\!\left(-\frac{z^2}{2\sigma_{\task}}\right)\,dz.
\]
If $f$ grows at most polynomially, then for $\sigma_{\task}$ in a neighbourhood of 1 the integrand is dominated by a function of the form $B(1+|z|^{4s})e^{-tz^2}$, with $t$ in a small neighbourhood of $1/2$. Thus the fourth moment is finite and, by dominated convergence, $\E[f^4(\sqrt{\sigma_{\task}}g)] \to \E[f^4(g)]$. The same argument applies to the other moments involving polynomials of $f$ and $g$ that are used below. Since $\E[\sigma_{\task}]=1$ and $\mathrm{Var}(\sigma_{\task})\to0$ by the assumptions above, we have $\sigma_{\task} \to 1$ in probability.

Given the above arguments, we will heuristically proceed in Appendix \ref{sec:pop_cond_avg_proof} with the usual treatment of $f(\x^\top\task)$ as $f(g)$ for $g\sim \mathcal{N}(0,1)$. Again, we reiterate that this replacement is true asymptotically using the trace scaling, effective rank, and uniform integrability assumptions above; it is not true that $\x^\top \task \overset{d}{=} g$ identically, the way it would if we assumed $\task$ was spherical, as is standard in the SIM literature{}

\newpage
\section{Population averages}
\label{sec:pop_cond_avg_proof}
From the assumptions and discussion in Appendix \ref{sec:gaussian_universality_general_covariance}, we will treat  $\task^\top\x\sim g$ asymptotically and proceed with the moment calculations using Hermite expansion. When $\|\task\|^2$ appears in an expression, we will write this as $d\omega$ for precision, knowing that $\omega$ is a random variable we are asymptotically treating as 1. 

\begin{boxedlemma}{Conditional expectations of order-two quantities}
For fixed $\|\task\|^2 = d\omega$ with $\omega \approx 1$, we have conditional expectations (over $\x$ and $y$) given by 
\begin{align}
  \E[y_i^2]
  &\approx\rho_f,
  \label{eq:exp_y2}\\
  \E[y_{\mathrm{test}}\vv{H}^{\,\top}]
  &\approx
  \frac1d
  \vv{
    \begin{bmatrix}
      a_1^2\task\task^\top/ \omega^2
      &
      a _1\rho_f\task/\omega
    \end{bmatrix}
  }^{\!\top},
  \label{eq:exp_yH}\\
  \E[\vv{H}\vv{H}^{\,\top}]
  &\approx
  \frac1d I_d\otimes
  \begin{bmatrix}
    \displaystyle\frac d\ell\rho_fI_d+a_1^2\task\task^\top/\omega^2
    &
    a_1\rho_f\task/\omega\\
    a_1\rho_f\task^\top/\omega
    &
    \rho_f^2
  \end{bmatrix}.
  \label{eq:exp_H2}
\end{align}
\end{boxedlemma}

\begin{proof}
First, we have 
\begin{align*}
  \E[f(g)] =\E[1\cdot f(g)]
    =\E\left[\sum_{m=0}^{\infty}a_mh_m(g)h_0(g)\right] =\sum_{m=0}^{\infty}a_m\E[h_m(g)h_0(g)]
    =a_0.
\end{align*}
Then \eqref{eq:exp_y2} follows from
\begin{align*}
  \E[y_i^2]
  &=\E[f^2(\task^\top\x)+2\varepsilon f(\task^\top\x)+\varepsilon^2]\\
  &\approx\E[f^2(g)+2\varepsilon f(g)+\varepsilon^2]\\
  &=\E f^2(g)+2\E\varepsilon\,\E f(g)+\E\varepsilon^2\\
  &=\rho+
    \E\left[
      \sum_{m=0}^{\infty}\sum_{p=0}^{\infty}
      a_ma_ph_m(g)h_p(g)
    \right] =\rho+\sum_{m=0}^{\infty}a_m^2.
\end{align*}

Decompose $\x$ into the component aligned with $\task$ and its orthogonal component:
\begin{align*}
  \E[y_i\x_i]
  &=
  \E\left[
    \left(
      \frac{\task\task^\top}{\|\task\|^2}\x_i+\x_{i,\perp}
    \right)
    f(\task^\top\x_i)
  \right]\\
  &=
  \frac{\task}{\|\task\|^2}
  \E[(\task^\top\x_i)f(\task^\top\x_i)]
  \approx
  \frac{\task}{d\omega }\E[gf(g)]\\
  &=
  \frac{\task}{d\omega}
  \sum_{m=0}^{\infty}a_m\E[h_1(g)h_m(g)]
  =
  \frac{a_1\task}{d\omega}.
\end{align*}

Combining these expressions,
\begin{align*}
  \E[y_{\mathrm{test}}H]
  &=
  \E\left[
    y_{\mathrm{test}}\x_{\mathrm{test}}
    \begin{bmatrix}
      \displaystyle\frac d\ell\sum_{i=1}^{\ell}y_i\x_i^\top
      &
      \displaystyle\frac1\ell\sum_{i=1}^{\ell}y_i^2
    \end{bmatrix}
  \right]\\
  &=
  \E[y_{\mathrm{test}}\x_{\mathrm{test}}]
  \begin{bmatrix}
    \displaystyle\E\left[\frac d\ell\sum_{i=1}^{\ell}y_i\x_i^\top\right]
    &
    \displaystyle\E\left[\frac1\ell\sum_{i=1}^{\ell}y_i^2\right]
  \end{bmatrix} \approx \frac{1}{d}
  \begin{bmatrix}
   \frac{a_1^2}{\omega^2} \task\task^\top & \frac{a_1 \rho_f}{\omega}\task
  \end{bmatrix}.
\end{align*}
Equation~\eqref{eq:exp_yH} follows by vectorising this expression. For the final moment calculation, we write
\begin{equation*}
    \E[\vv{H}\vv{H}^{\,\top}]
  =
  \frac{I_d}{d}\otimes
  \E\left[
    \begin{bmatrix}
      \bm u\bm u^\top&\bm u s\\
      s\bm u^\top&s^2
    \end{bmatrix}
  \right], \quad  \bm u=\frac d\ell\sum_i y_i\x_i,
  \quad s=\frac1\ell\sum_i y_i^2.
\end{equation*}

We compute each of the three blocks directly. 
\begin{align*}
\E[\bm{u}\bm{u}^\top] &=
\frac{d^2}{\ell^2}
\left(
  \sum_{i=1}^{\ell}\E[y_i^2\x_i\x_i^\top]
  +\sum_{i\ne j}\E[y_i\x_i]\E[y_j\x_j]^\top
\right)\\
&=
\frac{d^2}{\ell^2}
\sum_{i=1}^{\ell}
\left(
  \E[f^2(\task^\top\x_i)\x_i\x_i^\top]
  +\E[2\varepsilon f(\task^\top\x_i)\x_i\x_i^\top]
  +\E[\varepsilon^2\x_i\x_i^\top]
\right)
+\frac{d^2}{\ell^2}\ell(\ell-1)
\frac{a_1^2\task\task^\top}{d^2\omega^2}\\
&\approx
\frac{d^2}{\ell^2}
\sum_{i=1}^{\ell}
\left(
  \E[f^2(\task^\top\x_i)\x_i\x_i^\top]
  +\frac{\rho}{d}I_d
\right)
+\frac{(\ell-1)a_1^2}{\ell\omega^2}\task\task^\top\\
&=
\frac{(\ell-1)a_1^2}{\ell\omega^2}\task\task^\top
+\frac{d^2}{\ell^2}
\sum_{i=1}^{\ell}
\left[
  \E\left[
    f^2(\task^\top\x_i)
    \left(
      \frac{\task\task^\top}{\|\task\|^2}\x_i+\x_{i,\perp}
    \right)
    \left(
      \frac{\task\task^\top}{\|\task\|^2}\x_i+\x_{i,\perp}
    \right)^\top
  \right]
  +\frac{\rho}{d}I_d
\right]\\
&\approx
\frac{(\ell-1)a_1^2}{\ell\omega^2}\task\task^\top
+\frac{d^2}{\ell^2}
\sum_{i=1}^{\ell}
\left[
  \E\left[\frac{g^2f^2(g)}{d^2\omega^2}\task\task^\top\right]
  +\E[f^2(g)]\E[\x_{i,\perp}\x_{i,\perp}^\top]
  +\frac{\rho}{d}I_d
\right]\\
&=
\frac{(\ell-1)a_1^2}{\ell\omega^2}\task\task^\top
+\frac{\E[g^2f^2(g)]}{\ell\omega^2}\task\task^\top + \frac{d\sum_{m=0}^{\infty}a_m^2}{\ell}
\left(I_d-\frac{\task\task^\top}{\|\task\|^2}\right)
+\frac{d\rho}{\ell}I_d\\
&\approx
\frac d\ell
\left(\rho+\sum_{m=0}^{\infty}a_m^2\right)I_d
+\frac{a_1^2}{\omega^2}\task\task^\top + \frac{1}{\ell\omega^2}
\left(
  -a_1^2+\E[g^2f^2(g)]
  -(\rho_f-\rho)
\right)\task\task^\top.
\end{align*}
We neglect the last $1/\ell$ in our final expression. For the vector block, we have 
\begin{align*}
\E[\bm{u}s] &= 
\frac d{\ell^2}
\left(
  \sum_{i=1}^{\ell}\E[y_i^3\x_i]
  +\sum_{i\ne j}\E[y_i\x_i]\E[y_j^2]
\right)\\
&\approx
\frac d{\ell^2}
\left[
  \sum_{i=1}^{\ell}
  \E[f^3(g)\x_i+3\varepsilon^2f(g)\x_i]
  +\sum_{i\ne j}
  \frac{a_1\task}{d\omega}
  \left(\rho+\sum_{m=0}^{\infty}a_m^2\right)
\right]\\
&=
\frac1{\ell^2}
\left[
  \sum_{i=1}^{\ell}\frac{\task}{\omega}
  \left(\E[gf^3(g)]+3\rho a_1\right)
  +\sum_{i\ne j}\frac{a_1\task}{\omega}
  \left(\rho+\sum_{m=0}^{\infty}a_m^2\right)
\right]\\
&=
\frac{a_1\task}{\omega}
\left(\rho+\sum_{m=0}^{\infty}a_m^2\right)
+ \frac{\task}{\ell\omega}
\left[
  \E[gf^3(g)]+3\rho a_1
  -a_1\left(\rho+\sum_{m=0}^{\infty}a_m^2\right)
\right].
\end{align*}
Finally the scalar block is
\begin{align*}
\E\left[s^2
\right]
&=
\frac1{\ell^2}
\left(
  \sum_{i=1}^{\ell}\E y_i^4
  +\sum_{i\ne j}\E y_i^2\,\E y_j^2
\right)\\
&=
\frac1\ell
\left(
  \E f^4(g)
  +6\rho\sum_{m=0}^{\infty}a_m^2
  +3\rho^2
\right)
+\frac{\ell-1}{\ell}
\left(
  \rho+\sum_{m=0}^{\infty}a_m^2
\right)^2\\
&=
\left(
  \rho+\sum_{m=0}^{\infty}a_m^2
\right)^2 + \frac1\ell
\left[
  \E f^4(g)
  +6\rho\sum_{m=0}^{\infty}a_m^2
  +3\rho^2
  -\left(
    \rho+\sum_{m=0}^{\infty}a_m^2
  \right)^2
\right].
\end{align*}
Together, the leading terms in these three blocks give \eqref{eq:exp_H2}.
\end{proof}

\newpage
\section{Derivation of the ICL formula}
\label{sec:icl_formula_proof}
This section will demonstrate how our stated formula for ICL error follows directly from the moment calculations in Section \ref{sec:pop_cond_avg_proof} and the results and derivation of \cite{letey2025pretrain}. In particular, comparing equations \eqref{eq:exp_y2}, \eqref{eq:exp_yH}, and \eqref{eq:exp_H2} with Lemma 2 of \cite{letey2025pretrain}, we see that the nonlinear link function $f$ affects the second-moment expressions for $y^2$, $yH$, and $\mathrm{vec}(H)\mathrm{vec}(H)^\top$ only through the transformations
\begin{align*}
    C_\mathrm{train} &\to a_1^2 C_\mathrm{train} \\
    C_\mathrm{test} &\to a_1^2 C_\mathrm{test} \\
    \rho &\to \rho_\mathrm{nonlinear}.
\end{align*}
This immediately gives the stated result so long as $a_1 \neq 0$. 

To treat the case $a_1=0$, we return to the task-conditional second moments. Setting $a_1=0$ gives
\begin{align*}
    \E[y_i^2] &\to \rho_\mathrm{nonlinear} \\
    \E[y_{\mathrm{test}}H] &\to 0 \\
    \E[\mathrm{vec}(H)\mathrm{vec}(H)^\top] &\to \frac{1}{d}I_d \otimes \begin{bmatrix}
        \frac{1}{\alpha}\rho_\mathrm{nonlinear} I_d & \bm{0} \\
        \bm{0}^\top & \rho^2_\mathrm{nonlinear}
    \end{bmatrix}.
\end{align*} We see that there is no longer any task randomness in these moments, which simplifies the resolvent calculation. Following closely the calculation of \cite{letey2025pretrain}, we arrive at a similar self-consistent asymptotic expression for the extended resolvent \begin{align*}
    \mathcal{G}(\pi) &\approx \left(\frac{\tau}{1 + \chi_\pi}\begin{bmatrix}
        \rho_\mathrm{nonlinear} & 0 \\
        0 & I_d \otimes B
    \end{bmatrix} + \pi \begin{bmatrix}
        0 & 0 \\
        0 & I_d \otimes B 
    \end{bmatrix} + \tau\lambda I\right)^{-1} \\
    \chi_\pi &= \frac{1}{d}\tr\left(\left(\frac{\tau}{1+\chi_\pi}B + \pi B + \tau\lambda I\right)^{-1}B\right)
\end{align*}
for \begin{align*}
B = \begin{bmatrix}
        \frac{1}{\alpha}\rho_\mathrm{nonlinear} I_d & \bm{0} \\
        \bm{0}^\top & \rho^2_\mathrm{nonlinear}
    \end{bmatrix}.
\end{align*}
Previously, $B$ would have contained task randomness that would have made $\chi_\pi$ difficult to analyse. Here, however, we obtain the simpler equation
\begin{align*}
    \chi_0 \approx \left(\frac{\rho_\mathrm{nonlinear}}{\alpha}\frac{\tau}{1 + \chi_0} + \tau \lambda \right)^{-1}\left(\frac{\rho_\mathrm{nonlinear}}{\alpha}\right).
\end{align*} Write $\tilde{\lambda} = \lambda(1+\chi_0)$; we seek solutions where $\tilde{\lambda}$ is nonnegative. Taking the ridgeless limit $\lambda \to 0$, we obtain
$$\tilde{\lambda} = \begin{cases}
    \frac{\rho_\mathrm{nonlinear}}{\alpha}\frac{1-\tau}{\tau} & \tau < 1 \\
    0 & \tau > 1
\end{cases}.$$

Substituting $\tilde{\lambda}$ above and $C\to0$ directly into the formula of \cite{letey2025pretrain} gives
\begin{equation}
  \mathcal E_{\mathrm{ICL}}
  \to
  \rho_\mathrm{nonlinear}
  \begin{cases}
    \displaystyle\frac{1}{1-\tau}, & \tau<1, \\[2mm]
    \displaystyle\frac{\tau}{\tau-1}, & \tau>1,
  \end{cases}
\end{equation}
which is exactly \eqref{eq:no-linear-mode}. 

\end{document}